%% file: paper.tex
\documentclass[]{TEAI}
\usepackage{helvet}

\usepackage{amsmath} 
\usepackage{natbib}
\usepackage{graphicx}
\usepackage{subcaption} 

\usepackage[toc,page,header]{appendix}
\usepackage[utf8]{inputenc} 
\usepackage[T1]{fontenc}    
\usepackage{hyperref}       
\usepackage{url}            
\usepackage{booktabs}       
\usepackage{lmodern}        
\usepackage{amsfonts}       
\usepackage{nicefrac}       
\usepackage{microtype}      
\usepackage{wrapfig}

\usepackage{amssymb}  
\usepackage{fontawesome}  
\usepackage{url}  

\usepackage{titletoc}

\usepackage{tikz}  
\usepackage{comment}  
\usepackage{tabularx}  
\usepackage{booktabs}  

\usepackage{minitoc}

\usepackage{booktabs}
\usepackage{array}
\usepackage{etoolbox}

\definecolor{lightblue}{RGB}{200, 230, 255}  
\definecolor{headerblue}{RGB}{150, 200, 255} 

\usepackage{pgfplots}
\usepackage[utf8]{inputenc} 
\usepackage[T1]{fontenc}    
\usepackage{hyperref}       
\usepackage{url}            
\usepackage{booktabs}       
\usepackage{amsfonts}       
\usepackage{nicefrac}       
\usepackage{microtype}      
\usepackage{xcolor}         
\usepackage{graphicx}
\usepackage{float}
\usepackage{comment}
\usepackage{multirow} 
\usepackage{amsmath} 
\usepackage{makecell} 
\usepackage{siunitx}  
\usepackage{tikz}
\usepackage{pgf-pie} 
\usepackage{subcaption}
\usepackage{wrapfig}
\usepackage[export]{adjustbox}

\usepackage{ragged2e}      
\usepackage{tabularx}       
\usepackage{array}          
\usepackage{caption}        
\usepackage{enumitem}
\usepackage{pifont}
\usepackage[hang,flushmargin]{footmisc} 

\usepackage{tcolorbox}

\usepackage{tcolorbox}    
\tcbuselibrary{breakable}  
\tcbuselibrary{skins}      

\usepackage{tabularx}
\usepackage{listings}

\title{\textsc{DCDM}: Divide-and-Conquer Diffusion Models for Consistency-Preserving Video Generation}

\author{
    Haoyu Zhao\textsuperscript{1,*},
    Yuang Zhang\textsuperscript{2},
    Junqi Cheng\textsuperscript{2},  
    Jiaxi Gu\textsuperscript{2},
    Zenghui Lu\textsuperscript{2},\\
    Peng Shu\textsuperscript{2},
    Zuxuan Wu\textsuperscript{1,$\dagger$},
    Yu-Gang Jiang\textsuperscript{1,$\dagger$}
}

\affiliation[1]{\mbox{Fudan University}} 
\affiliation[2]{\mbox{Tencent}}

\abstract{
\begin{abstract}

Recent video generative models have demonstrated impressive visual fidelity, yet they often struggle with semantic, geometric, and identity consistency. In this paper, we propose a system-level framework, termed the \textsc{Divide-and-Conquer Diffusion Model (DCDM)}, to address three key challenges: (1) intra-clip world knowledge consistency, (2) inter-clip camera consistency, and (3) inter-shot element consistency. DCDM decomposes video consistency modeling under these scenarios into three dedicated components while sharing a unified video generation backbone. For intra-clip consistency, DCDM leverages a large language model to parse input prompts into structured semantic representations, which are subsequently translated into coherent video content by a diffusion transformer. 
For inter-clip camera consistency, we propose a temporal camera representation in the noise space that enables precise and stable camera motion control, along with a text-to-image initialization mechanism to further enhance controllability.
For inter-shot consistency, DCDM adopts a holistic scene generation paradigm with windowed cross-attention and sparse inter-shot self-attention, ensuring long-range narrative coherence while maintaining computational efficiency. We validate our framework on the test set of the CVM Competition at AAAI'26, and the results demonstrate that the proposed strategies effectively address these challenges.
\end{abstract}
}

\correspondence{\email{zxwu@fudan.edu.cn} and \email{ygj@fudan.edu.cn}}
\checkdata[Website]{\href{https://sites.google.com/view/aaai26-cvm/home/}{AAAI 2026 Consistency in Video Generative Models (CVM) Competition}}
\checkdata[Team]{Cinematrix}
\checkdata[Award]{CVM @ AAAI'26 Top Team}

\begin{document}
\maketitle
\renewcommand{\thefootnote}{}
\footnotetext{$^*$This work was completed during an internship at Tencent under the Qingyun Project.\\$^\dagger$Corresponding authors.}
\renewcommand{\thefootnote}{\arabic{footnote}}


\vspace{-1.5em}

\input{section/introduction}

\input{section/related}
\input{section/method}
\input{section/results}

\input{section/conclusion}

\clearpage

\bibliographystyle{plainnat}
\bibliography{main}






\end{document}

%% file: section/introduction.tex
\section{Introduction}

Recent advances in video generative models have significantly improved visual fidelity and short-term temporal coherence. However, achieving consistency across diverse generation scenarios remains a fundamental challenge. In practice, video consistency manifests along multiple dimensions, including intra-clip world knowledge consistency, inter-clip camera consistency, and inter-shot element consistency.
These dimensions pose inherently different modeling requirements. Intra-clip consistency demands grounded semantic reasoning to ensure that objects, attributes, and physical interactions within a single clip remain plausible and non-contradictory. Inter-clip camera consistency requires explicit modeling of camera motion and view transitions, which are temporally structured and fundamentally geometric in nature. Inter-shot element consistency, in contrast, focuses on long-range narrative coherence, requiring persistent memory of characters, scenes, and styles across multiple shots and extended time horizons.

Due to these fundamentally distinct sources of inconsistency, enforcing all three forms of consistency within a single monolithic video diffusion model is highly challenging. A unified model must simultaneously reason over semantics, geometry, and long-range temporal dependencies, often leading to representational entanglement and suboptimal trade-offs between controllability, coherence, and computational efficiency. As a result, existing approaches tend to excel in one dimension while compromising others.

In this paper, we propose a system-level framework termed the Divide-and-Conquer Diffusion Model (DCDM). Rather than enforcing all consistency constraints within a single monolithic diffusion model, DCDM decomposes the problem into three dedicated components, each explicitly designed to address one key dimension of consistency, while sharing a common video diffusion backbone.
Specifically, for intra-clip world knowledge consistency, our goal is to ensure that the generated video remains semantically and physically coherent within a single clip. To this end, we first employ a large language model to parse the input prompt, extracting the implicit subjects, attributes, and scene descriptions embedded in natural language. These structured semantic representations are then provided to a diffusion transformer (DiT)~\cite{peebles2023scalable}, which translates the parsed world knowledge into visually coherent video content. This design allows DCDM to ground video generation in explicit semantic reasoning, reducing contradictions and implausible interactions within a clip.

To achieve inter-clip camera consistency, DCDM explicitly models camera motion as a first-class control signal. We again leverage a language model to analyze the prompt and classify the described camera motion into predefined motion categories. Based on this analysis, we propose a temporally structured noise representation that is injected into the diffusion process to guide camera motion across frames.
Furthermore, to enhance controllability and stability, we adopt a text-to-image initialization strategy, where a reference image is first generated from the textual description. The diffusion model then conditions on both the reference image and the camera representation, enabling camera-controlled video generation with improved temporal coherence and precise motion execution.

For inter-shot element consistency, we focus on maintaining the identity and attributes of characters, scenes, and styles across long-form narratives. We employ a holistic scene generation model that produces the entire video sequence jointly, ensuring global coherence from the first shot to the last. 
To enable fine-grained directorial control, our architecture incorporates a window cross-attention mechanism to localize textual prompts to specific shots, while a sparse inter-shot self-attention pattern, which is dense within shots but sparse across shots, balances long-range consistency with the computational efficiency required for minute-scale generation.
Beyond improving narrative coherence, this design exhibits strong emergent properties, including persistent character and scene memory, as well as an implicit understanding of cinematic structures and shot composition.

In conclusion, DCDM demonstrates that video consistency is best addressed through a divide-and-conquer strategy, where semantic reasoning, camera control, and narrative coherence are modeled in a disentangled yet unified diffusion framework. The main contributions of this work are summarized as follows:

\begin{itemize}
    \item We propose the Divide-and-Conquer Diffusion Model (DCDM), a unified system framework that decomposes video consistency into three complementary dimensions, including intra-clip world knowledge consistency, inter-clip camera consistency, and inter-shot element consistency.
    
    \item For intra-clip world knowledge consistency, we introduce an LLM-based prompt extension mechanism that enriches implicit semantics to ensure coherent video generation within each clip.
    
    \item For inter-clip camera consistency, we propose to control the camera movement in noise space with a temporal structure that enables precise and stable camera motion control in diffusion models.
    
    \item For inter-shot element consistency, we introduce sparse inter-shot self-attention that enables efficient long-range coherence while preserving dense intra-shot modeling.
\end{itemize}

%% file: section/related.tex
\section{Related Work}

Recent advances in diffusion models based on U-Net and Diffusion Transformer (DiT)~\cite{peebles2023scalable} architectures have significantly improved video generation~\cite{wang2025uniadapter,chen2024videocrafter2,zhao2025magdiff,yang2024cogvideox,wan2025wan,hacohen2024ltx,zhao2026lstd,zhao2025dynamictrl}. U-Net–based approaches employ an encoder–decoder structure that offers efficient inference and architectural flexibility, but their limited model capacity constrains performance gains under large-scale scaling. As a result, recent methods, such as CogVideoX~\cite{yang2024cogvideox}, Seedance~\cite{gao2025seedance}, Seaweed-7B~\cite{seawead2025seaweed}, and Wan-14B model~\cite{wan2025wan}, increasingly adopt DiT as the core building block, leveraging stacked Transformer layers to jointly model spatial and temporal dependencies, with scalable depth that better supports large-scale training.
These works encompass a broad spectrum of diffusion-based architectures, from U-Net variants to Transformer-centric designs, coupled with advances in temporal modeling, conditional control, sampling efficiency, and hierarchical synthesis. These developments collectively push the frontier toward more controllable and high-fidelity video generation.
Despite these advances, modeling long-term dependencies and preserving consistency in complex scenes remains challenging due to the interplay of high-dimensional spatial detail and rich temporal dynamics.

%% file: section/method.tex
\section{Method}

\begin{figure*}[t!]
    \centering
    \includegraphics[width=0.90\linewidth]{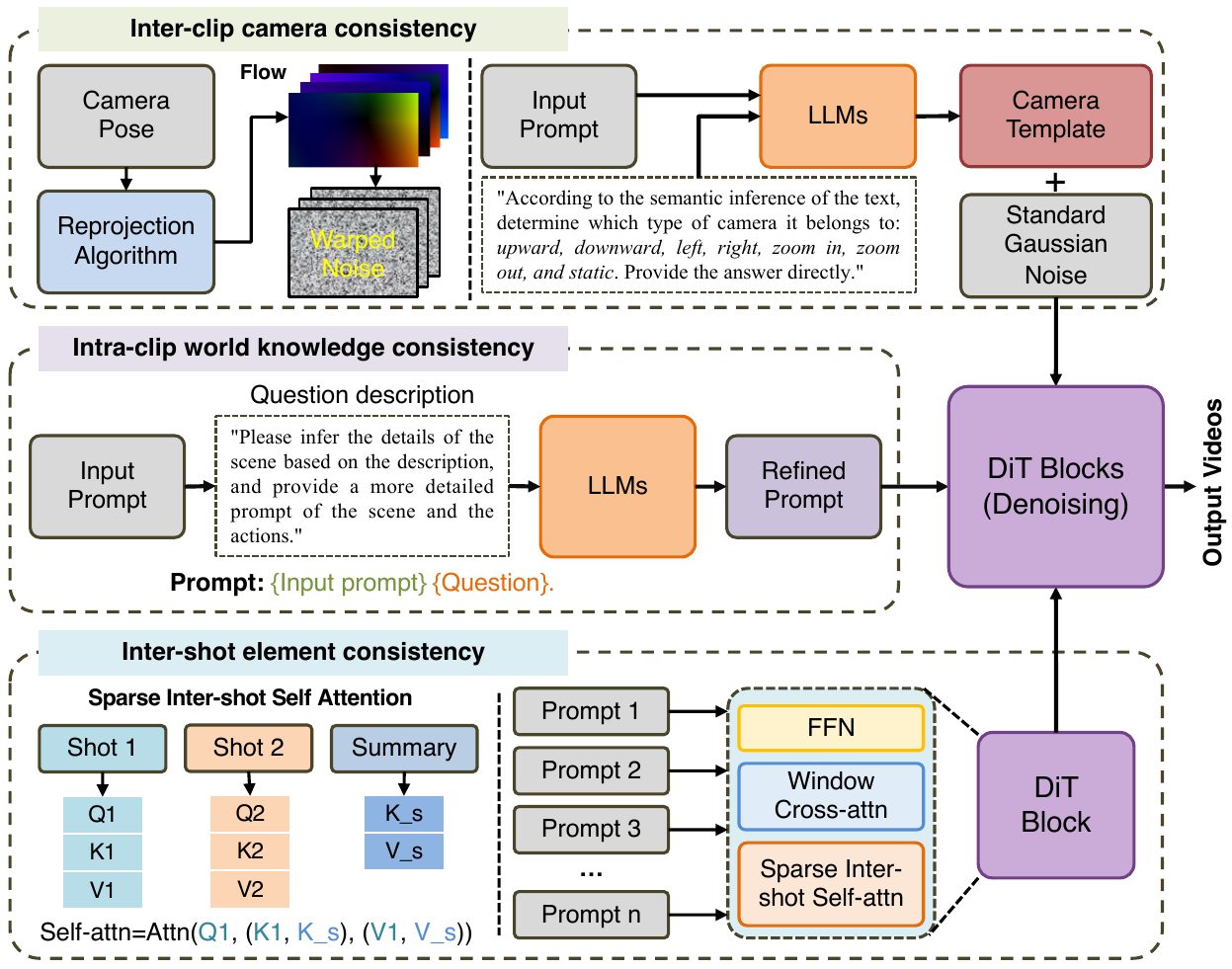}
    \caption{A high-level overview of our proposed Divide-and-Conquer Diffusion Model (DCDM) framework.}
    \label{fig:intro}
\end{figure*}

We propose the \textbf{Divide-and-Conquer Diffusion Model (DCDM)}, a system-level framework that addresses video consistency by decomposing it into three complementary dimensions: (i) intra-clip world knowledge consistency, (ii) inter-clip camera consistency, and (iii) inter-shot element consistency. As illustrated in Fig.~\ref{fig:intro}, all components share a unified video diffusion backbone based on Diffusion Transformers~\cite{wan2025wan}, while each consistency dimension is handled by a dedicated module with task-specific inductive biases. This design avoids forcing heterogeneous consistency constraints into a single monolithic model and enables scalable, controllable video generation across diverse scenarios.

\subsection{Preliminaries}
\label{sec:preliminaries}

We adopt a latent video diffusion formulation. Given a video latent $\mathbf{x}_0 \in \mathbb{R}^{T \times H \times W \times C}$, a forward diffusion process gradually corrupts it with Gaussian noise. A DiT-based denoiser $\epsilon_\theta$ is trained to predict the injected noise at diffusion step $t$:
\begin{equation}
\mathcal{L} = \mathbb{E}_{\mathbf{x}_0, t, \epsilon}\left[
\|\epsilon - \epsilon_\theta(\mathbf{x}_t, t \mid \mathbf{c})\|_2^2
\right],
\end{equation}
where $\mathbf{c}$ denotes conditioning signals. In DCDM, $\mathbf{c}$ is constructed differently depending on the target consistency dimension, while the DiT backbone remains shared.

\subsection{Intra-clip World Knowledge Consistency}
\label{sec:intra_clip}

\paragraph{Motivation.}
Intra-clip consistency requires that objects, attributes, actions, and physical interactions within a single clip remain semantically and physically coherent. However, user prompts are often under-specified, leaving critical world knowledge implicit. Relying on a diffusion model to infer such knowledge implicitly often leads to semantic contradictions or implausible dynamics.

\paragraph{LLM-based Prompt Extension.}
To address this issue, we introduce an explicit semantic Prompt Extension (PE) stage using a large language model (\textit{e.g.,} Qwen3~\cite{yang2025qwen3}). Given an input prompt $p$, we pair it with a fixed instruction that requests detailed inference of scene content and actions. The LLM produces a refined prompt $\tilde{p}$:
\begin{equation}
\tilde{p} = \mathrm{LLM}(p, \text{question}),
\end{equation}
where $\tilde{p}$ expands implicit information into explicit descriptions of entities, attributes, spatial relations, and plausible actions.

The refined prompt $\tilde{p}$ is then embedded and provided to the DiT backbone as textual conditioning:

\begin{equation}
\mathbf{c}_{\text{text}} = E_{\text{text}}(\tilde{p}).
\end{equation}

By separating semantic reasoning from visual generation, this module significantly improves intra-clip world knowledge consistency without modifying the diffusion architecture.

\subsection{Inter-clip Camera Consistency via Camera Control}
\label{sec:camera_noise}

\paragraph{Motivation.}
Camera motion consistency is inherently geometric and temporally structured. Conditioning camera behavior solely through text often yields weak or unstable control. 
We therefore introduce a temporally coherent noise representation that is injected into the initialization stage of diffusion sampling using the given camera parameters.
Given an input prompt, we employ Qwen3 to classify the described camera motion into a predefined category set $\mathcal{M}$, including ``left, right, upward, downward, zoom in, zoom out, static'', and so on. The predicted motion category $m \in \mathcal{M}$ determines a corresponding camera template $\mathcal{T}_m$, which specifies the canonical temporal motion pattern.

\paragraph{Camera Representation Construction}
We construct the camera representation in noise space by warping Gaussian noise across time according to camera motion. Starting from standard Gaussian noise at the first frame:

\begin{equation}
\mathbf{z}_1 \sim \mathcal{N}(0, \mathbf{I}).
\end{equation}

Then, we propagate it using a reprojection operator $\mathcal{W}_{t \leftarrow t-1}$ derived from camera intrinsics and extrinsics:

\begin{equation}
\tilde{\mathbf{z}}_t = \mathcal{W}_{t \leftarrow t-1}(\mathbf{z}_{t-1}).
\end{equation}

To preserve stochasticity and avoid over-correlated noise, we blend the warped noise with independent Gaussian noise:

\begin{equation}
\mathbf{z}_t = \sqrt{\lambda}\,\tilde{\mathbf{z}}_t + \sqrt{1-\lambda}\,\epsilon_t, \quad \epsilon_t \sim \mathcal{N}\left(0, \mathbf{I}\right),
\end{equation}

where $\lambda$ controls temporal coherence strength. The warping schedule and parameters are specified by $\mathcal{T}_m$.

\paragraph{Noise-space Injection}
Unlike step-wise conditioning, the proposed camera representation is injected only at the initialization stage of diffusion sampling by replacing the initial noise:

\begin{equation}
\mathbf{x}_T \leftarrow \{\mathbf{z}_1, \dots, \mathbf{z}_T\}.
\end{equation}

This anchors the entire denoising trajectory to a temporally coherent noise structure, yielding stable and precise camera motion without interfering with later diffusion dynamics.

\paragraph{Reference Image Conditioning}
To further enhance controllability, we optionally generate a reference image $I_{\text{ref}}$ using a text-to-image model (Z-Image~\cite{cai2025z}) and condition the DiT backbone on both $I_{\text{ref}}$ and the camera representation. This reduces appearance ambiguity and improves camera execution.

\begin{figure*}[t!]
    \centering
    \includegraphics[width=1.0\linewidth]{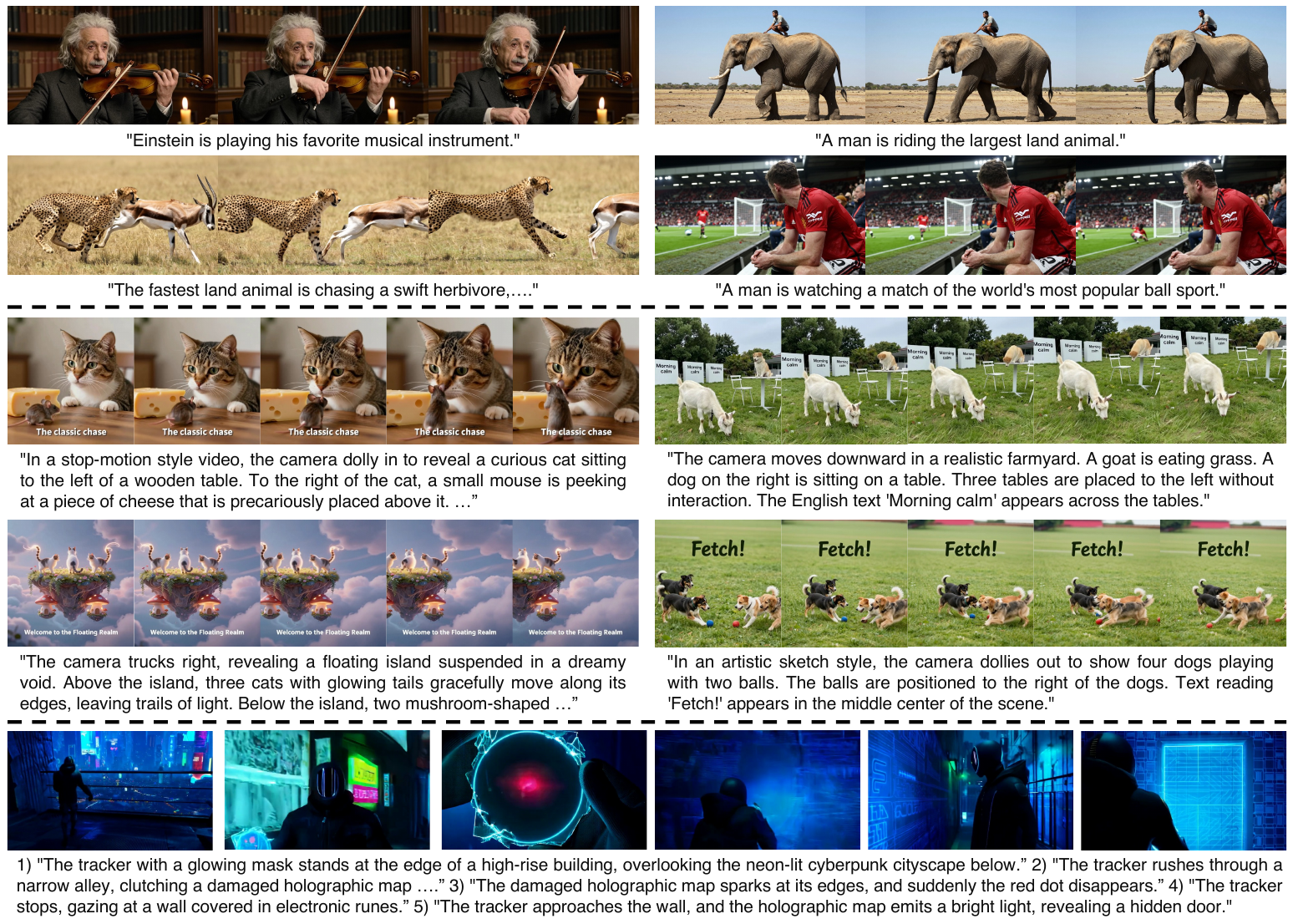}
    \caption{Qualitative results of our DCDM framework across intra-clip world knowledge consistency (above), inter-clip camera consistency (middle), and inter-shot element consistency scenarios (bottom).}
    \label{fig:results}
\end{figure*}

\subsection{Inter-shot Element Consistency with Sparse Inter-shot Self-Attention}
\label{sec:inter_shot}

\paragraph{Motivation.}
Inter-shot consistency focuses on preserving character identity, environment, and style across multiple shots. Crucially, intra-shot and inter-shot consistency exhibit different properties: intra-shot consistency requires dense temporal modeling, while inter-shot consistency primarily requires selective information exchange. Full attention across all frames is unnecessary and computationally prohibitive.

\paragraph{Sparse Inter-shot Self-Attention.}
Following the method~\cite{meng2025holocine}, we employ a sparse inter-shot self-attention mechanism that significantly reduces complexity while preserving essential information flow.
For each shot $i$, we apply full bidirectional self-attention among its tokens. Queries $\mathbf{Q}_i$ attend to keys and values $\mathbf{K}_i, \mathbf{V}_i$ within the same shot, ensuring smooth motion and local coherence.
To enable information exchange across shots, we construct a global context summary. For each shot $j$, we select a small representative subset of key-value tokens $\mathbf{KV}_{\text{summary},j}$ (e.g., tokens from the first frame). These summaries are concatenated to form a global cache:

\begin{equation}
\mathbf{KV}_{\text{global}} = \left[\mathbf{KV}_{\text{summary},1}, \dots, \mathbf{KV}_{\text{summary},N_s}\right],
\end{equation}

where $N_s$ is the number of shots. Queries from shot $i$ attend to both local and global keys:

\begin{equation}
\mathrm{Attn}\left(\mathbf{Q}_i, \mathbf{KV}\right) = \mathrm{Attn}\left(\mathbf{Q}_i, \left[\mathbf{KV}_{\text{global}}, \mathbf{K}_i, \mathbf{V}_i\right]\right).
\end{equation}

\paragraph{Complexity Analysis.}
For a video with $N_s$ shots, each of length $L_{\text{shot}}$, full attention incurs $O\left(\left(N_s L_{\text{shot}}\right)^2\right)$ complexity. In contrast, our sparse attention reduces complexity to approximately $O\left(N_s \cdot L_{\text{shot}} \cdot S\right)$, where $S$ is the number of summary tokens per shot and $S \ll L_{\text{shot}}$.

\paragraph{Integration into DiT.}
Finally, we integrate this design into DiT blocks by combining (i) windowed cross-attention for shot-specific prompt localization and (ii) sparse inter-shot self-attention. This enables efficient minute-scale video generation with strong narrative coherence.

%% file: section/results.tex
\section{Results}

\paragraph{Qualitative Results}
As illustrated in Fig.~\ref{fig:results}, we demonstrate the effectiveness of DCDM across the three targeted video generation scenarios. For these diverse tasks, our model consistently achieves high-quality video generation with strong temporal coherence and continuity.

%% file: section/conclusion.tex
\section{Conclusion}

In this paper, we presented the Divide-and-Conquer Diffusion Model (DCDM), a unified framework for addressing key consistency challenges in video generation. By decomposing the problem into intra-clip world knowledge consistency, inter-clip camera consistency, and inter-shot element consistency, DCDM provides targeted solutions through structured semantic parsing, camera signal control, and efficient attention mechanisms. Experiments on the AAAI'26 CVM Competition benchmark demonstrate that the proposed framework effectively improves semantic coherence, camera stability, and narrative continuity. These results confirm the potential of system-level modeling for reliable and controllable video synthesis.